\documentclass[runningheads]{llncs}
\usepackage{graphicx}
\usepackage{comment}
\usepackage{hyper ref}

\usepackage{amsmath}
\usepackage{amssymb,url,natbib}
\usepackage{dsfont}
\usepackage{caption}
\usepackage{subcaption}
\usepackage{wrapfig}
\usepackage{color}
\usepackage{xcolor}
\definecolor{Gray}{gray}{0.9}
\definecolor{Orange}{rgb}{0.86, 0.52, 0.0}
\definecolor{black}{rgb}{0.0, 0.0, 0.0}

\usepackage{multirow}
\usepackage{colortbl}
\usepackage{array}
\usepackage{todonotes}
\usepackage{tablefootnote}
\usepackage{footnote}
\makesavenoteenv{tabular}
\makesavenoteenv{table}

\usepackage{algorithm}
\usepackage{algorithmic}

\usepackage{float}
\usepackage{ulem}

\usepackage[symbol]{footmisc}

\newcommand{\beginsupplement}{%
        \setcounter{page}{1}%
        \setcounter{table}{0}
        \renewcommand{\thetable}{S\arabic{table}}%
        \setcounter{figure}{0}
        \renewcommand{\thefigure}{S\arabic{figure}}%
     }

\newcolumntype{g}{>{\columncolor{Gray}}l}
\newcolumntype{G}{>{\columncolor{Gray}}c}

\begin{document}
\title{Fed-MIWAE: Federated Imputation of Incomplete Data via Deep Generative Models}
\titlerunning{Fed-MIWAE: Federated Imputation of Incomplete Data}
%


\author{Anonymous} 

\author{Irene Balelli\inst{1\dag}\orcidID{0000-0002-4593-8217}
\and
Aude Sportisse\inst{1\dag}\orcidID{0000-0003-1303-0762} \and
 Francesco Cremonesi\inst{1}\orcidID{0000-0003-1027-485X} \and
 Pierre-Alexandre Mattei\inst{1}\orcidID{0000-0002-1297-908X} \and
 Marco Lorenzi\inst{1}\orcidID{0000-0003-0521-2881} for the  Alzheimer’s  Disease  Neuroimaging  Initiative\thanks{Data  used  in  preparation  of  this  article  were  obtained  from  the  Alzheimer’s  Disease Neuroimaging  Initiative  (ADNI)  database  (\url{adni.loni.usc.edu}).  As  such,  the  investigators within the ADNI contributed to the design and implementation of ADNI and/or provided data but  did  not  participate  in  analysis  or  writing  of  this  report.  A  complete  listing  of  ADNI investigators can be found at: \url{http://adni.loni.usc.edu/wp-content/uploads/how_to_apply/ADNI_Acknowledgement_List.pdf}}
 }

\renewcommand{\thefootnote}{\fnsymbol{footnote}}
\footnotetext[2]{The two authors contributed  equally to this paper.}
\renewcommand{\thefootnote}{\arabic{footnote}}

\authorrunning{I. Balelli, A. Sportisse et al.}

 \institute{Centre Inria d'Universit\'e C\^ote d'Azur \\
 \email{\{irene.balelli,aude.sportisse,francesco.cremonesi,pierre-alexandre.mattei,marco.lorenzi\}@inria.fr}}
%
\maketitle              
\begin{abstract}
Federated learning allows for the training of machine learning models on multiple decentralized local datasets without requiring explicit data exchange. However, data pre-processing, including strategies for handling missing data, remains a major bottleneck in real-world federated learning deployment, and is typically performed locally. This approach may be biased, since the subpopulations locally observed at each center may not be representative of the overall one. To address this issue, this paper first proposes a more consistent approach to data standardization through a federated model. Additionally, we propose Fed-MIWAE, a federated version of the state-of-the-art imputation method MIWAE, a 
deep latent variable model for missing data imputation 
based on variational autoencoders. MIWAE has the great advantage of being easily trainable with classical federated aggregators. Furthermore, it is able to deal with MAR (Missing At Random) data, a more challenging missing-data mechanism than MCAR (Missing Completely At Random), where the missingness of a variable can depend on the observed ones. We evaluate our method on multi-modal medical imaging data and clinical scores from a simulated federated scenario with the ADNI dataset. We compare Fed-MIWAE with respect to classical imputation methods, either performed locally or in a centralized fashion. Fed-MIWAE allows to achieve imputation accuracy 
comparable with the best centralized method, even when local data distributions are highly heterogeneous. In addition, 
thanks to the variational nature of Fed-MIWAE, our method is designed to perform multiple imputation, allowing for the quantification of the imputation uncertainty in the federated scenario.
\end{abstract}

\section{Introduction and previous work}

Digital health revolution relies heavily on the massive development and deployment of data-driven machine learning (ML) and deep learning (DL) methods. 
Nevertheless, in order to insure reliability, trustworthiness and generalizability of such models, their parameters need to be learned from a sufficiently large curated dataset, representative of the whole data distribution. These conditions are rarely met within a single research hospital, and data centralization is often not possible due to current privacy regulations (\textit{e.g.} the European GDPR\footnote{https://gdpr-info.eu/}). 

Federated learning (FL) is an emerging learning paradigm based on multiple decentralized datasets. FL has been identified as a promising solution for healthcare applications \citep{rieke2020future,crowson2022systematic}, due to its improved robustness and generalization properties. Indeed, the final aggregated FL model gathers information provided by multiple hospitals through distributed optimization. Moreover, FL implicitly alleviates data governance burden and naturally improves privacy guarantees with respect to classical centralized ML methods. 
The growing interest in FL is demonstrated by the several FL-based research projects~\citep{sadilek2021privacy,crowson2022systematic,rieke2020future,pati2022federated}, and FL infrastructures such as Flare \citep{roth2022flare}, OpenFL~\citep{foley2022openfl}, SubstraFL\footnote{https://docs.substra.org/en/stable/substrafl\_doc/substrafl\_overview.html}, and Fed-BioMed\footnote{https://fedbiomed.gitlabpages.inria.fr}, which are specifically tailored for biomedical applications. 
Despite the potential of FL for cross-institutional medical research, data pre-processing, including data standardization and  management of missing entries in the database, is generally carried out locally by each participating site. The importance of establishing reliable data pre-processing pipelines in a federated manner has already been emphasized \citep{rajula2019overview}, and is of paramount importance in real-life FL deployment. 

In order to prevent the deletion of incomplete rows in the dataset, with consequent critical loss of information (especially in the high-dimensional setting) and the potential induction of a biased analysis, one common approach to handle missing values involves data imputation \citep{little2019statistical}. Besides the naive mean imputation, which consists 
in replacing missing values with the mean across the observed population, 
several alternatives and more elaborate imputation methods have been proposed in the literature, including based on probabilistic models \citep{little2019statistical}, low-rank models \citep{mazumder2010spectral} or random forests \citep{stekhoven2012missforest}. Some methods are specifically designed for multiple imputation, such as \textsc{mice}
\citep{van2011mice}, to account for imputation uncertainty  \cite{austin2021missing}.  More recently, imputation methods based on deep latent variable models have been proposed, for example by building on  generative
adversarial networks (GANs) \citep{li2018learning,yoon2018gain} or on  variational autoencoders (VAEs) \citep{mattei2019miwae,nazabal2020handling,ipsen2020not}. 
However, none of these imputations methods are specifically designed for the FL setting. 
The theoretical behaviour of learning with hidden variables in a federated context has been addressed in \cite{NEURIPS2021_f740c8d9} through an Expectation Maximization (EM) based optimization scheme, and applied to federated missing values imputation for biodiversity monitoring. In \citep{balelli2021probabilistic}, authors proposed a Bayesian latent variable model for multiple views data assimilation and completely missing modalities imputation in a federated setting \citep{balelli2021probabilistic}, which also relies on the EM algorithm for local parameters' optimization. 
Recently, two papers \citep{zhou2021federated,yao2022fedtmi} have extended the imputation method based on GANs \citep{yoon2018gain} to the FL framework. 
Nevertheless, the theoretical results behind these methods have only been provided in the Missing Completely At Random (MCAR) setting \citep{yoon2018gain}, and are not suited for the more complex and yet relevant case of Missing At Random (MAR) data, where the missingness of a variable depends on the observed ones, or the most challenging Missing Not At Random (MNAR) setting, where the missingness can even depend on the unobserved variable itself \citep{mohan2018handling,austin2021missing} (Table \ref{tab:missing_type}). It is worth mentioning that 
it is not possible to assess whether the missing pattern is MCAR or MAR solely based on the data (\textit{i.e.} without additional clinical knowledge), hence the need in favouring methods able to handle both cases, and possibly explicitly account for the uncertainty on the imputed variable, for instance through multiple imputation.
\begin{table}[h!]
\caption{Missing data classification with examples}
\label{tab:missing_type}
\centering
\begin{tabular}{l|p{11cm}}
\textbf{MCAR}  &  Accidental loss/damage of a sample, non-systematic personnel error in a sample's acquisition/reporting. \\
\hline 
\textbf{MAR}  &  The acquisition of a feature (\textit{e.g.} a clinical score test) depends on a fully observed variable (\textit{e.g.} the age): the observed systematic difference in incomplete features can be explained by differences in observed variables. \\
\hline
\textbf{MNAR}  &   The reason for a missing sample can depend both on observed and unobserved variables, including the missing sample itself (\textit{e.g.} self-censored data)
\end{tabular}
\end{table}

In this work, we first propose a pipeline 
for data pre-processing through multi-site data, including federated standardization as well as imputation of missing entries, performed through Fed-MIWAE, the federated version of MIWAE~\citep{mattei2019miwae}, a deep latent variable model for missing data imputation based on variational inference. 
The VAEs-based structure of MIWAE makes it easily trainable with classical federated aggregators, such as FedAvg~\citep{mcmahan2017communication}, FedProx~\citep{li2018federated} or  Scaffold~\citep{karimireddy2020scaffold}. 
By relying on the framework of MIWAE,  Fed-MIWAE is naturally designed to handle both MCAR and MAR data, while allowing for multiple imputation. 
In Sec. \ref{sec:method} we describe our proposed approach and its implementation, 
and in Sec. \ref{sec:results} we applied it to multi-modal medical imaging data and clinical scores from the ADNI database, distributed to multiple sites to simulate a federated setting. Fed-MIWAE shows higher generalization and improved performance compared to all tested local imputation methods, and comparable with the centralized scenario, even in presence of simulated MAR data and high heterogenity accross local datasets in terms of patients distribution. 

\section{An approach to federated data pre-processing}\label{sec:method}

\subsection{Missing data importance-weighted
autoencoder (MIWAE)}

In this work we extend the MIWAE method \citep{mattei2019miwae} in order to be able to train it in a federated manner. MIWAE is defined by a deep generative model 
from a latent Gaussian variable $z$ to the original data $x$, where $t \sim p_\theta(x|z)$, and the conditional distribution $p_\theta$ is parametrized using a neural network (NN) with weights $\theta$.
Due to the presence of missing values in the data, the observed likelihood we would like to maximise is intractable  
\citep{little2019statistical}. Similarly to variational autoencoders, the idea of MIWAE is to propose a conditional distribution of the latent variables given the observed data,  $x^{\textrm{observed}}$, 
$q_\gamma(z|x^\textrm{observed})$,
 parameterized as well by a NN with weights $\gamma$. A stochastic lower bound of the observed likelihood can then be computed and optimized using SGD-like approaches. 
Finally, the imputation step is done by computing the conditional expectation of the missing variables given the observed ones, $\mathbb{E}[x^\textrm{missing}|x^\textrm{observed}]$,  estimated via self-normalised importance sampling.

%

\subsection{Proposed method}


We propose a federated pre-processing approach,  described in Algorithm \ref{alg:second}. 
Firstly, we perform federated standardization in presence of missing values through a single communication round (Stage 1). This step only requires the communication to the central server of all local feature-wise means and standard deviations to analytically evaluate the global mean and std. Each center will be able to use this information to perform its data standardization with respect to the whole federated dataset. Secondly, we propose to train Fed-MIWAE using the same FL platform (Stage 2 (i) of Algorithm \ref{alg:second}), over the decentralised standardised datasets, across $R$ iterative communication rounds. 
In this stage, the local parameters' optimization and aggregation can be performed using classical FL optimization and aggregation schemes, such as Federated Averaging (FedAvg) \citep{mcmahan2017communication}, FedProx \citep{li2020federated} or Scaffold \citep{karimireddy2020scaffold}, which are designed to be more robust to data heterogeneity. 
Finally, missing values can be imputed locally by each center using the final global model (Stage 2 (ii) of Algorithm \ref{alg:second}).

    \begin{algorithm}[tb]
	\caption{Federated data pre-processing}
	\label{alg:second}
	\begin{algorithmic}
	\small
            \STATE {\bfseries Input:} Rounds $R$, Epochs $E$, MIWAE hyperparameters 
		\STATE {\bfseries \underline{Stage 1: Federated standardization}} 
            \STATE Each center $c$ computes its local mean and std $(\boldsymbol\mu^{(c)},\boldsymbol\sigma^{(c)})$.
            \STATE Server aggregates them and sends the global mean and std $(\boldsymbol\mu,\boldsymbol\sigma)$.
            \STATE Each center $c$ standardizes its local dataset with $\boldsymbol\mu$ and $\boldsymbol\sigma$.
            \STATE {\bfseries \underline{Stage 2: Fed-MIWAE}} 
            \STATE {\bfseries (i) \textit{Train step}}
		\FOR{$r=1$ {\bfseries to} R}
            \STATE Each center $c$ initializes its local model with $(\theta^{r-1},\gamma^{r-1})$.
            \STATE Each center performs $E$ epochs of local MIWAE bound optimization, and sends the updated local parameters $((\theta^{(c)})^{t+1},(\gamma^{(c)})^{t+1})$ to the Server.
            \STATE Server aggregates $((\theta^{(c)})^{t+1},(\gamma^{(c)})^{t+1})_c$ and sends the global parameter $(\theta^{r},\gamma^{r})$ back to the centers.
		\ENDFOR
            \STATE {\bfseries (ii) \textit{Missing values imputation}}
            \STATE Each center $c$ dispose of the final global model $(\theta^R,\gamma^R)$ for missing data imputation.
	\end{algorithmic}
    \end{algorithm}



\section{Application to the Alzheimer’s Disease Neuroimaging Initiative dataset (ADNI)}\label{sec:results}



\subsection{Data}


We consider 311 participants extracted from the ADNI dataset,
 composed by 104 cognitively normal individuals (NC) and 207 patients diagnosed with Alzheimer disease (AD), coming from 55 centers. 
All participants are associated with multiple data views: cognitive scores including MMSE, CDR-SB, ADAS-Cog-11 and RAVLT, Magnetic resonance imaging (MRI), Fluorodeoxyglucose-PET and AV45-Amyloid PET images. MRI morphometrical biomarkers were obtained as regional volumes using the cross-sectional pipeline of FreeSurfer v6.0 based on the Desikan-Killiany parcellation~\citep{fischl2012freesurfer}. Measurements from AV45-PET and FDG-PET were estimated by co-registering each modality to their respective MRI, normalizing by the cerebellum uptake and by computing regional amyloid load and glucose hypometabolism using PetSurfer pipeline~\citep{greve2014cortical} based on the same parcellation. Features were corrected beforehand with respect to intra-cranial volume, sex and age using a multivariate linear model. Overall, the features dimension for a fully observed patient is 130. Further details are provided in Supp. Material, Table \ref{Tab_adni_demog}.

\subsection{Scenarios} 
\begin{figure}[ht]
     \begin{subfigure}[h]{0.5\textwidth}
         \includegraphics[width=\textwidth]{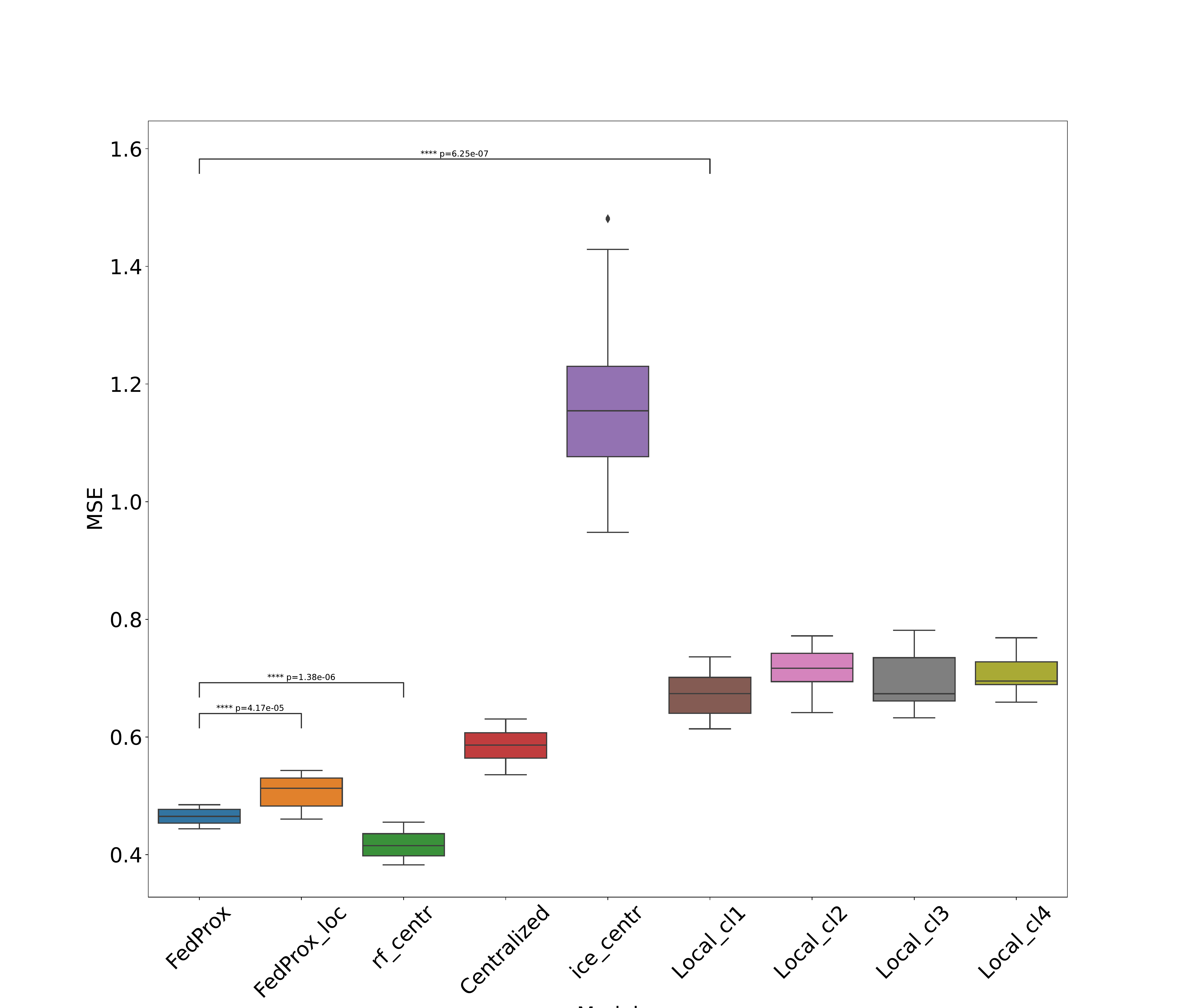}
         \caption{ }
         
     \end{subfigure}
     \hfill
     \begin{subfigure}[h]{0.5\textwidth}
         \includegraphics[width=\textwidth]{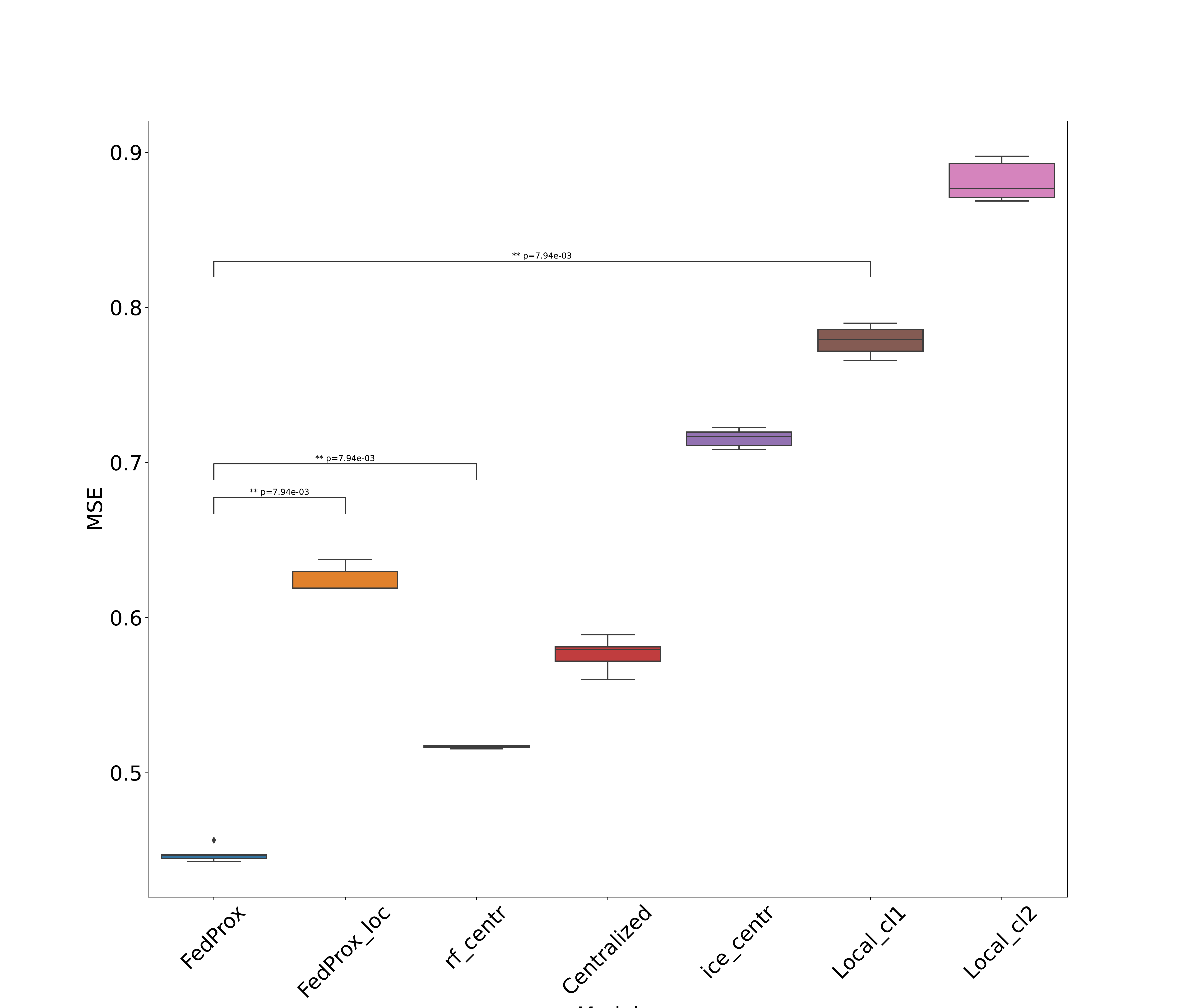}
         \caption{ }
         
     \end{subfigure}
        \label{Fig_MSE_ADNI_ext_MCAR}
   
     \begin{subfigure}[h]{0.5\textwidth}
         \includegraphics[width=\textwidth]{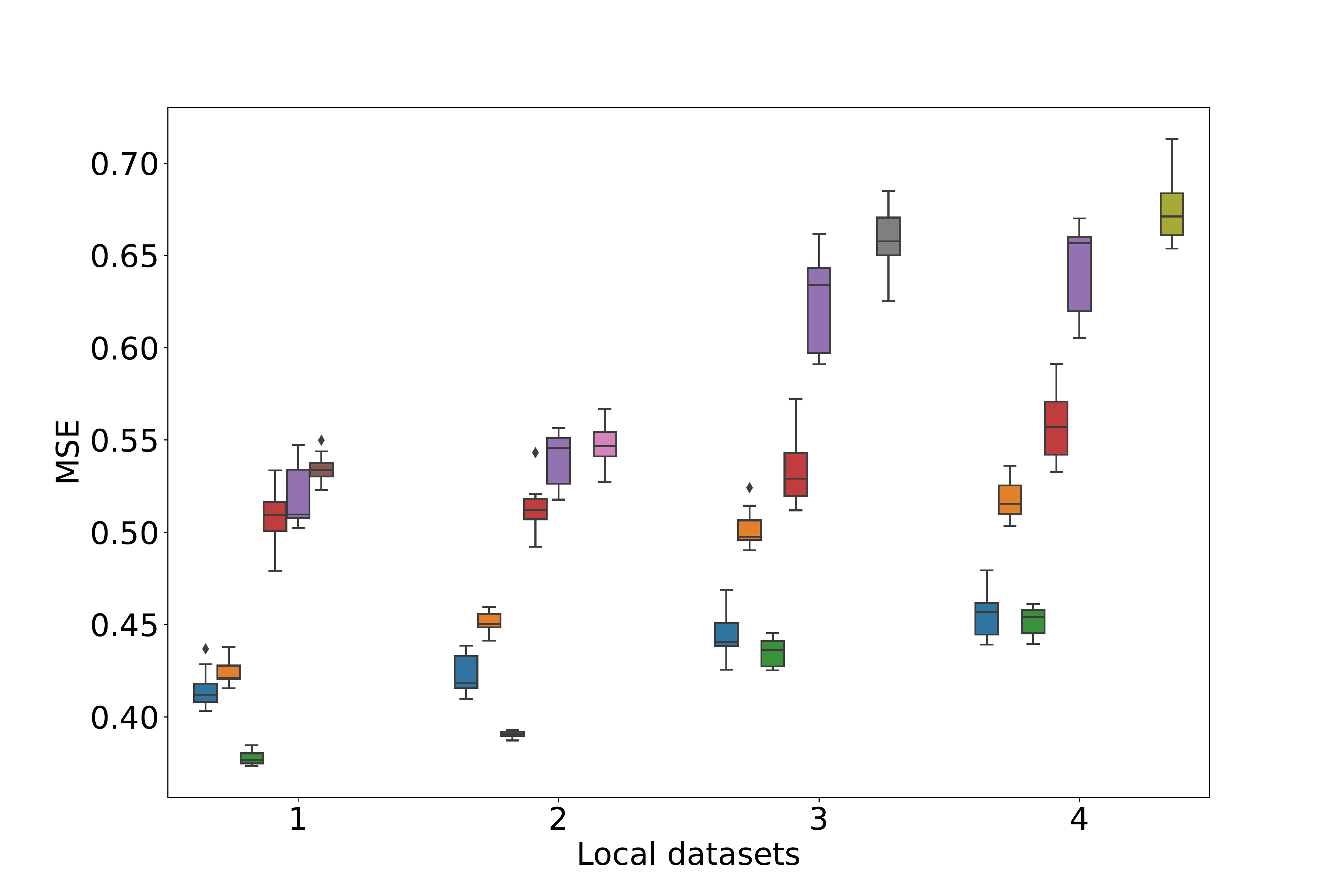}
         \caption{ }
          
     \end{subfigure}
     \hfill
     \begin{subfigure}[h]{0.5\textwidth}
         \includegraphics[width=\textwidth]{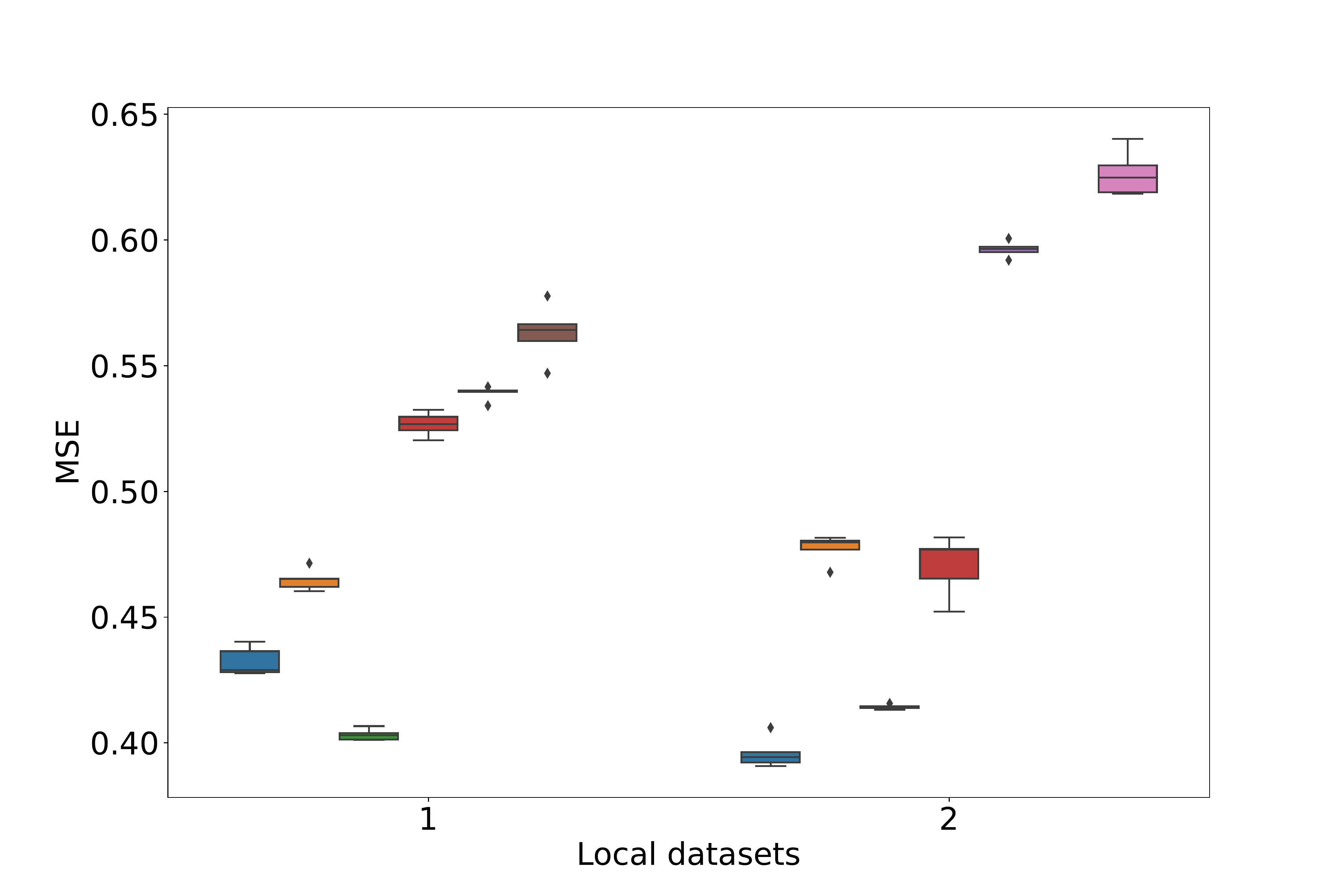}
         \caption{ }
       
     \end{subfigure}
        \caption{ADNI data, MCAR setting. Comparison of federated and local models measured on imputing missing entries of a previously unseen testing dataset (upper row) and of local datasets used for training (bottom row, where colors denotes the employed model, following the same color code as in the upper row figures). Left column: Natural split scenario. Right column: Not-IID scenario.}
        \label{Fig_MSE_ADNI_local_MCAR}
\end{figure}

We consider two federated scenarios:
\begin{description}
\item[\textbf{Natural split scenario}:] we take into account the information concerning the site to perform the split. Due to the high number of participating sites, we artificially create 4 datasets by gathering the contribution of about 14 ADNI sites per simulated center (see details in Supp. Material, Table \ref{Tab_adni_site_demog}). 
\item[\textbf{Not-IID scenario}:] we generated 3 centers, one containing only AD patients (Client 1, $N_1 = 171$), one containing only  NC patients (Client 2, $N_2 = 77$), and the third one, kept as external testing dataset, containing samples from both groups ($N_3 = 63$).
\end{description}
We further simulate 30$\%$ of missing data in each resulting local dataset, considering either the MCAR or MAR settings. For the natural split scenario, we perform 5 times a 4-fold cross-validation: each time, three datasets are used as the training centers, while the fourth one is used in testing phase only. For the not-IID scenario, the training is always performed (5 times) on Client 1 and Client 2, with testing on Client 3. 
Numerical experiments with Fed-MIWAE were executed 
using version v4.2 of the Fed-BioMed software\footnote{https://gitlab.inria.fr/fedbiomed/fedbiomed/-/tree/v4.2.4}. 
Experiment results were logged using Fed-BioMed's logging facility and manually collected for analysis.
Due to space limitations, we will only show results on the more challenging MAR setting in Sec. \ref{subsec:results}. Similar conclusions have been obtained in the MCAR setting, and are provided in the Supp. Materials (Table \ref{tab:notIID_MCAR_test}, 
Figure \ref{Fig_MSE_ADNI_local_MCAR}). 

\subsection{Benchmark}

We compare our proposed method trained using FedProx \citep{li2020federated} (simply denoted \texttt{FedProx} in Sec. \ref{subsec:results}, or \texttt{FedProx\_loc}, if the data standardization is done locally and only Stage 2 of Algorithm \ref{alg:second} is performed), 
with the local training of MIWAE in each site 
(\texttt{Local\_cl}$i$ for the site $i$) 
as well as the centralized training of MIWAE (\texttt{Centralized}). 
To ensure a fair comparaison, the chosen hyperparameters of MIWAE are the same, regardless of how the model is trained (Supp. Material, Table \ref{Tab_hyper_miwae}). In addition, we assess the performance of classical imputation methods, by training them either locally or using the centralized dataset:  naive mean imputation (\texttt{mean}),  nonparametric method with random forests (\texttt{rf}) \cite{stekhoven2012missforest} and iterative chained equations method (\texttt{ice}) \citep{van2011mice}, from the impute module of the scikit-learn library \cite{scikit-learn}. 
To evaluate the imputation accuracy of all considered methods, we compute the normalized Mean Squared Error (MSE).

\begin{figure}[h!]
     \begin{subfigure}[h]{0.55\textwidth}
         \includegraphics[width=\textwidth]{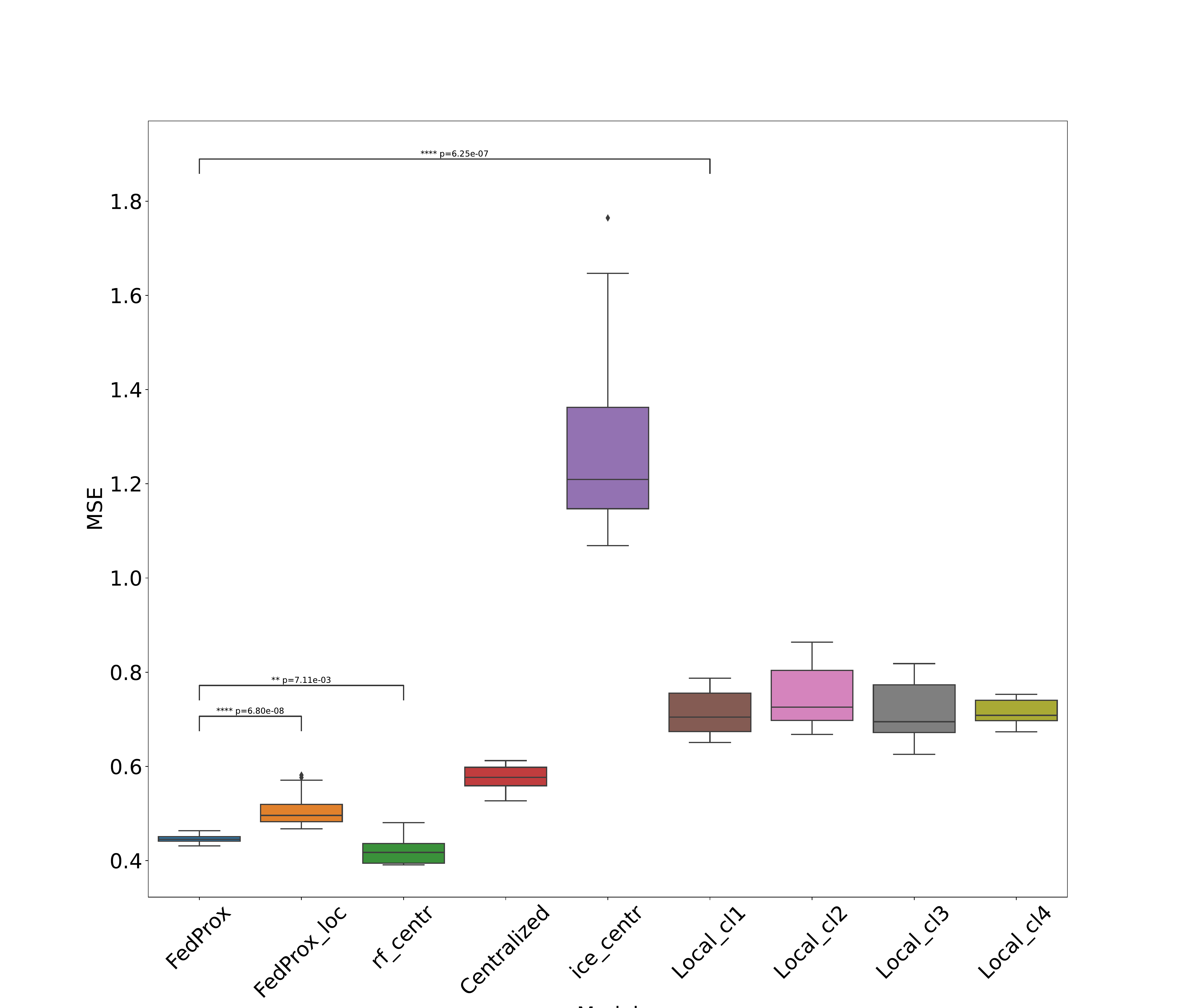}
         \caption{ }
         \label{Fig_MSE_ADNI_ext_natural}
     \end{subfigure}
     \hfill
     \begin{subfigure}[h]{0.55\textwidth}
         \includegraphics[width=\textwidth]{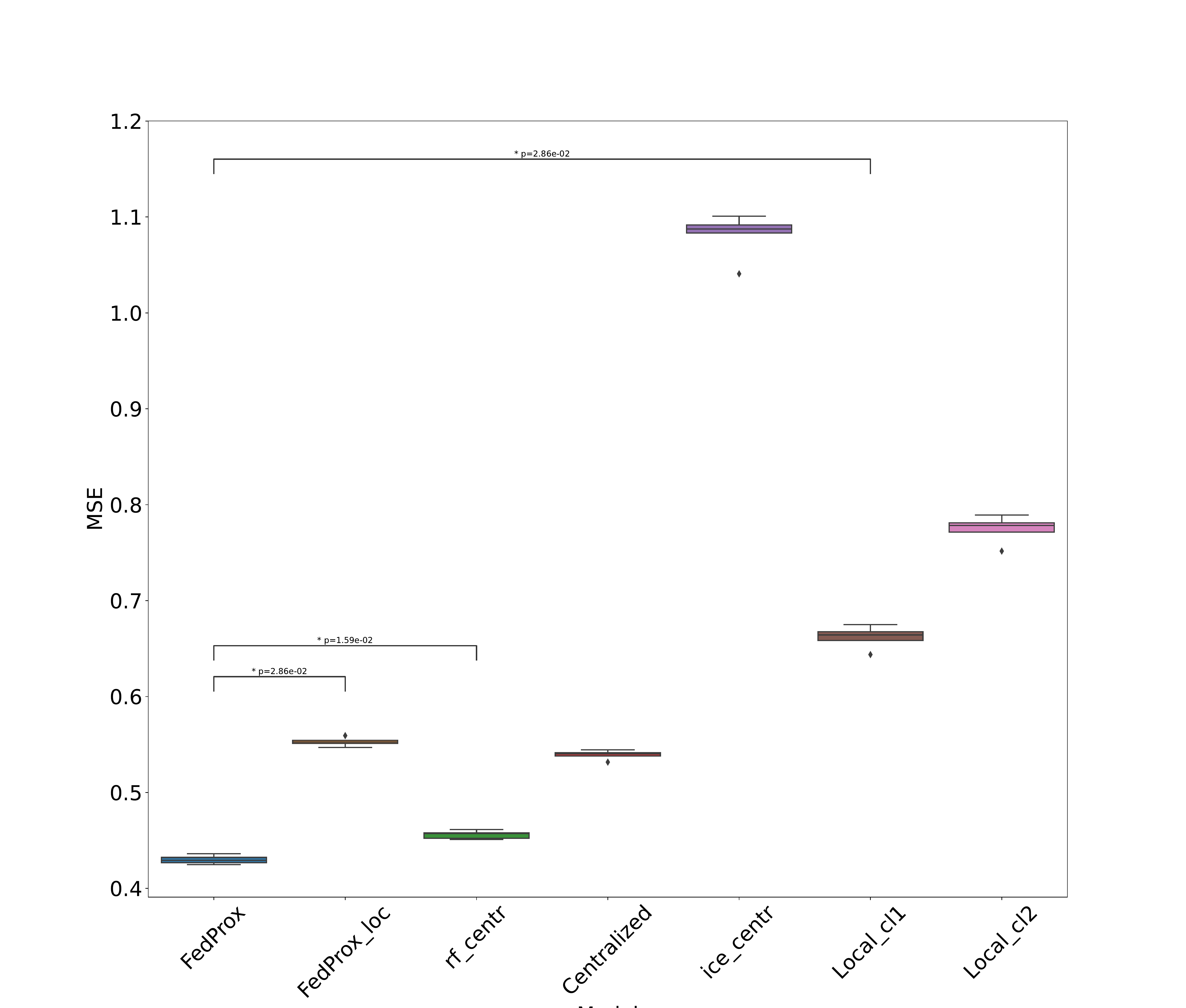}
         \caption{ }
         \label{Fig_MSE_ADNI_ext_notiid}
     \end{subfigure}
        \label{Fig_MSE_ADNI_ext}
     \begin{subfigure}[h]{0.55\textwidth}
         \includegraphics[width=\textwidth]{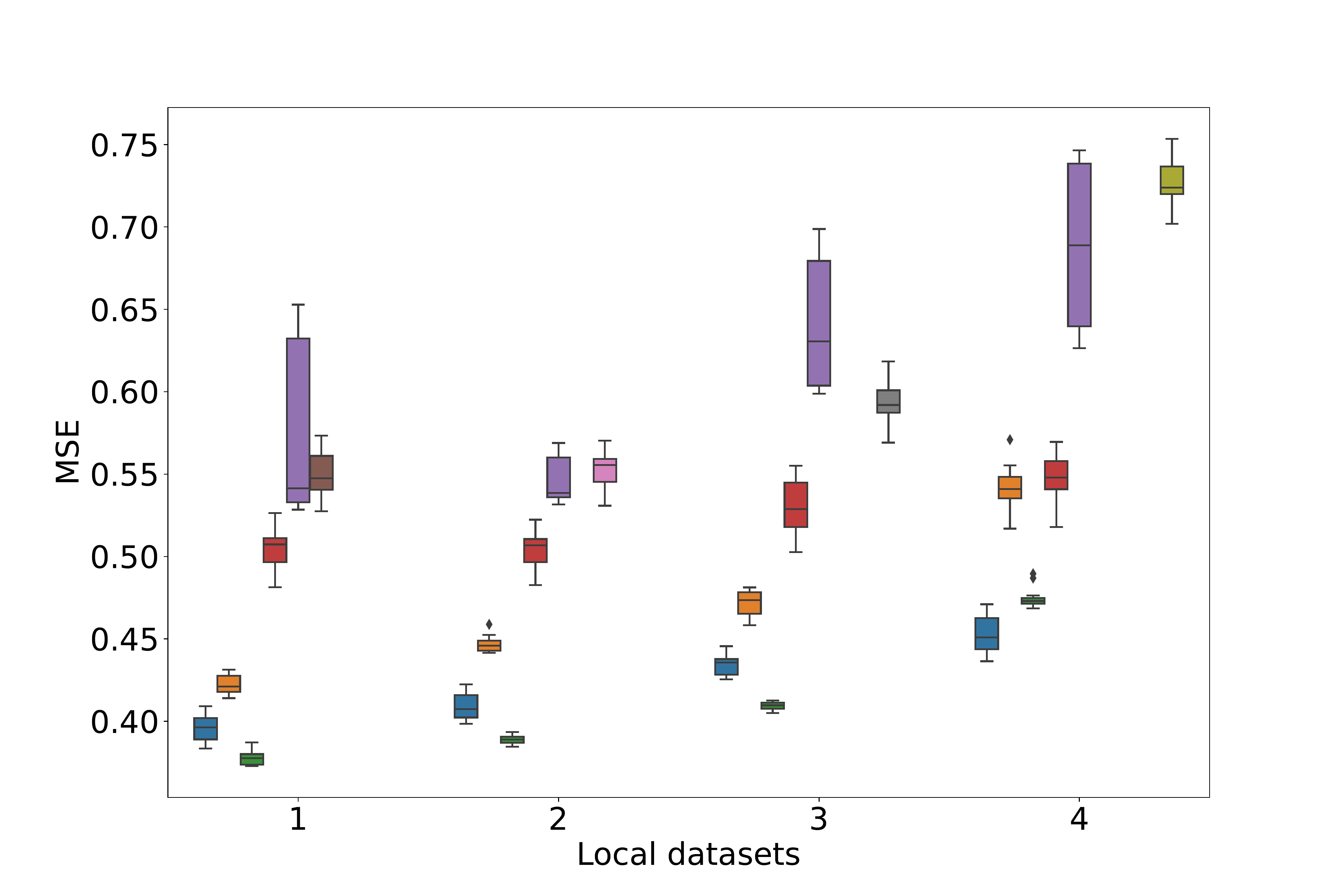}
         \caption{ }
          \label{Fig_MSE_ADNI_local_natural}
     \end{subfigure}
     \hfill
     \begin{subfigure}[h]{0.55\textwidth}
         \includegraphics[width=\textwidth]{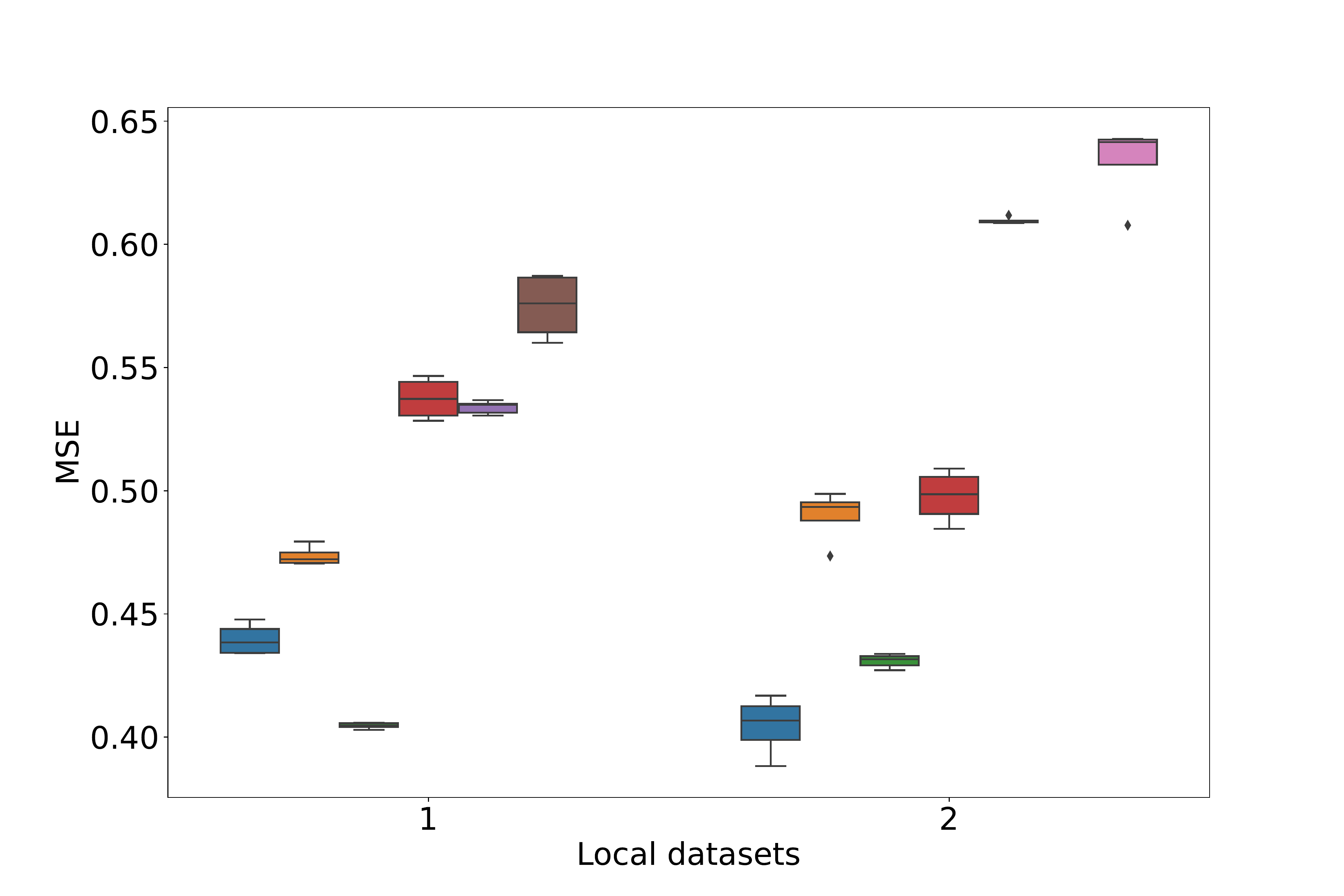}
         \caption{ }
         \label{Fig_MSE_ADNI_local_notiid}
     \end{subfigure}
        \caption{ADNI data, MAR setting. Comparison of federated and local models measured on imputing missing entries of a previously unseen testing dataset (upper row) and of local datasets used for training (bottom row, where colors denotes the employed model, following the same color code as in the upper row figures). Left column: Natural split scenario. Right column: Not-IID scenario. More details can be found in Table \ref{tab:notIID_MAR_test} and Supp. Table \ref{tab:natural_MAR_test}}
        \label{Fig_MSE_ADNI_MAR}
\end{figure}

\subsection{Results}\label{subsec:results}


In Figure \ref{Fig_MSE_ADNI_MAR}, we assess the performance of the imputation methods through MSE in both considered scenarios, with respect to the task of imputing missing data both on the external testing dataset (upper row) and on datasets which have been seen during training (bottom row). In all cases, our method optimized with FedProx outperforms all local MIWAE models (\texttt{Local\_cl}$i$). Moreover, one can see that performing data standardisation in a federated manner allows to reach improved results compared to the case where the standardization step has been performed locally (\textit{cf} \texttt{FedProx} compared to \texttt{FedProx\_loc}), especially in the more challenging non-iid scenario. Interestingly, the MIWAE method in the centralized setting displays a lower performance compared to Fed-MIWAE. This can be attributed to the potential issue of overfitting, and can be prevented for instance by deploying drop-out and early stopping strategies, which has not been done here for the sake of a fair comparison throughout the methods.


Among the considered benchmark approaches, one can see that the classical Random Forests imputation method (\texttt{rf}) achieves the highest performance, when tested on both an external dataset and the completion of missing entries from the training data. 
The naive mean-values imputation method \texttt{mean} is not included in Figure \ref{Fig_MSE_ADNI_MAR} due to its very poor performance (see for instance Table \ref{tab:notIID_MAR_test}), while the method \texttt{ice}, is significantly less accurate than \texttt{rf} and our proposed approach, despite being trained in a standard centralized manner. 

Fed-MIWAE shows a very good performance in all considered settings, compared to the centralized approaches tested. In particular, the imputation quality of our federated method is remarkable when considering the not-IID scenario and the ability of the trained model to infer missing entries in an external dataset (Figure \ref{Fig_MSE_ADNI_MAR} (d) and Table \ref{tab:notIID_MAR_test}), outperforming the best centralized model.
In addition, one can appreciate the consistent performance exhibited by Fed-MIWAE in the not-IID setting, where it achieves a MSE similar to the one obtained in the natural split scenario, where patients' distribution across centers is more balanced. In contrast, the performance of all local models (\texttt{Local\_cl}$i$) in imputing missing values of the external testing dataset drops significantly in this extremely non-heterogeneous scenario. Finally, our approach is even able to outperform the best centralized model when applied to the smaller sized local datasets (\textit{i.e.} Client 4 in the natural split scenario and Client 2 in the non-iid one), underscoring the importance of performing the imputation task by exploiting the federated framework.


\begin{table}[ht]
\caption{ADNI data, MAR setting, not-IID split. Comparison of all considered methods on the testing dataset (mean (std)).}\label{tab:notIID_MAR_test}
\centering
\begin{tabular}{l||g|g|l|l|l}
\textbf{Training} & \multicolumn{2}{G|}{Federated} & \multicolumn{3}{c}{Centralized} \\
\hline
\textbf{Model} & {\textbf{FedProx}} & FedProx\_loc & mean & ice & random forests \\
\hline
\textbf{MSE $\boldsymbol{\downarrow}$} & \textbf{0.4299 (0.005)} & 0.5529 (0.005) & 1.0933 & 1.081 (0.0233) & 0.4559 (0.0042)
\end{tabular}
\end{table}

\begin{figure}[ht]
     \begin{subfigure}[h]{0.4\textwidth}
         \includegraphics[width=\textwidth]{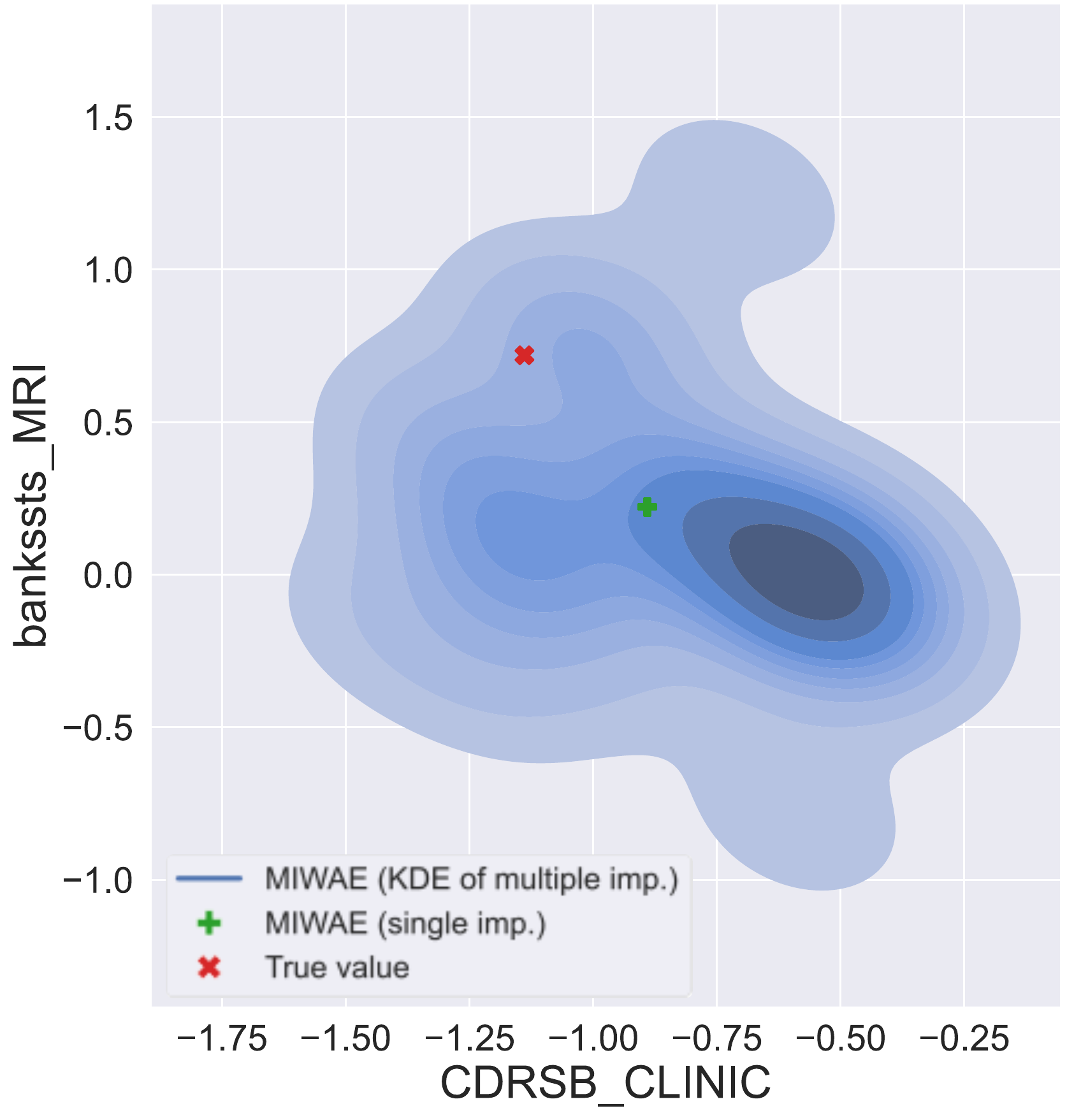}
         \caption{ }
     \end{subfigure}
     \hfill
     \begin{subfigure}[h]{0.4\textwidth}
         \includegraphics[width=\textwidth]{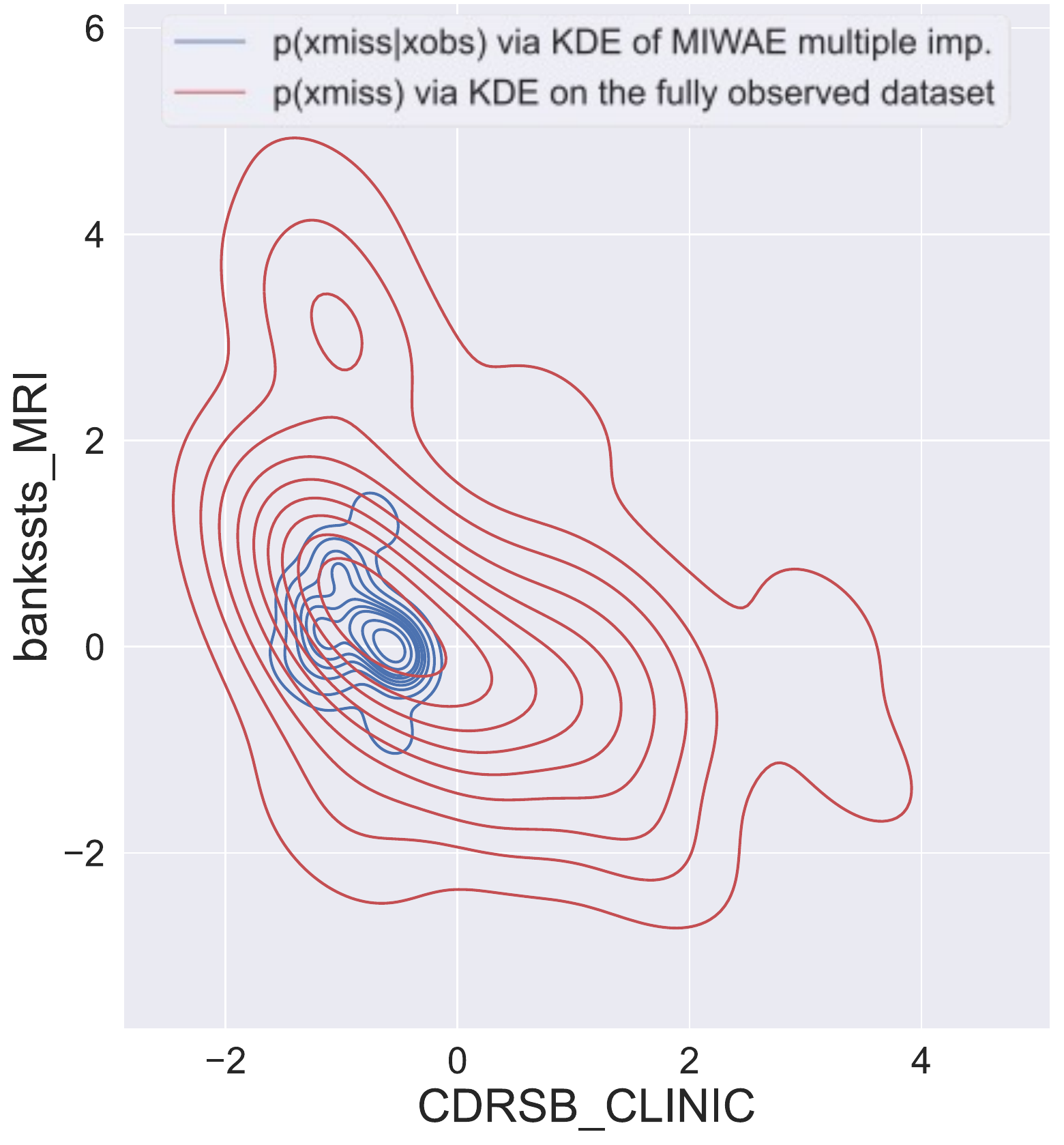}
         \caption{ }
     \end{subfigure}
        \caption{ADNI data, MAR setting. (a) Example of multiple imputation on the test data using Algorithm \ref{alg:second}, and (b) corresponding comparison between $p_{\textrm{Fed-MIWAE}}(x^{\textrm{missing}}|x^{\textrm{observed}})$ with $p(x^{\textrm{missing}})$.}
        \label{Fig_mult_imp_adni}
\end{figure}

Finally, we should stress that an additional major advantage of our method consists in its ability to perform multiple imputation, by drawing i.i.d. samples from the estimated posterior probability $p_{\textrm{Fed-MIWAE}}(x^{\textrm{missing}}|x^{\textrm{observed}})$. This allows us to quantify the uncertainty of the imputation. In Figure \ref{Fig_mult_imp_adni} (a) we show an example of multiple imputation on a randomly selected subject in the testing dataset obtained by drawing 30 samples, while in Figure \ref{Fig_mult_imp_adni} (b) one can see that a significantly lower variance is associated to $p_{\textrm{Fed-MIWAE}}(x^{\textrm{missing}}|x^{\textrm{observed}})$ compared to $p(x^{\textrm{missing}})$, leading us to the conclusion that $x^{\textrm{observed}}$ do provide relevant information for the imputation of $x^{\textrm{missing}}$.



\section{Conclusion}

In this work, we propose an efficient pipeline to handle MCAR and MAR values in the FL framework, which provides highly generalizable results and comparable performance with respect to centralized methods. We show that our method is accurate even when applied in highly heterogeneous settings, which should be carefully taken into account in real-world FL contexts. Several improvement of the proposed method can be envisaged in the future, spanning from addressing the complex problem of handling MNAR data,  by incorporating the not-MIWAE method  \citep{ipsen2020not} in our pipeline, to improve privacy requirements for instance by using secure aggregation or differential privacy techniques. In addition, data pre-processing pipelines in the federated framework should be further explored since they are key for the successful training of FL in real clinical studies.
\bibliographystyle{plainnat}
\bibliography{biblio}

\newpage
\section*{Supplementary Material}
\beginsupplement


\begin{table}[ht]
\centering
\caption{Demographics of the clinical sample from ADNI.}\label{Tab_adni_demog}
\centering
\begin{tabular}{llccc}
\textbf{Group} & \textbf{Sex} & \textbf{Count}  & \textbf{Age}   & \textbf{Range}\\ 
\hline 
\multirow{ 2}{*}{AD}   & Female  & 94     & 71.58 (7.59)    & 55.10 - 90.30   \\
   & Male         & 113    & 74.37 (7.19)     & 55.90 - 89.30    \\ 
   \hline
\multirow{ 2}{*}{NC}    & Female    & 58      & 73.76 (4.61)   & 65.10 - 84.70      \\
  & Male  & 46          & 75.39 (6.58)     & 59.90 - 85.60    
\end{tabular}
\end{table}

\begin{table}[ht]
\caption{Clinical sample's sizes per simulated center, natural split.}\label{Tab_adni_site_demog}
\centering
\begin{tabular}{lGGccGGcc}
\textbf{Center} & \multicolumn{2}{G}{1} & \multicolumn{2}{c}{2} & \multicolumn{2}{G}{3} & \multicolumn{2}{c}{4} \\
\hline
\multirow{2}{*}{\textbf{Count}} & AD & 57 & AD & 66 & AD & 41 & AD & 43  \\
& NC & 35 & NC & 38 & NC & 21 & NC & 10 \\
\hline
\textbf{Total} & \multicolumn{2}{G}{92} & \multicolumn{2}{c}{104} & \multicolumn{2}{G}{62} & \multicolumn{2}{c}{53}
%
\end{tabular}
\end{table}

\vspace{-.5cm}

\begin{table}[h!]
\caption{Fed-MIWAE Hyperparameters used. $C= $ Number of clients.}\label{Tab_hyper_miwae}
\centering
\begin{tabular}{l|c|l|p{1.3cm}|p{1.3cm}|p{2cm}|p{2cm}}
 \textbf{Training} & \textbf{Rounds} & \textbf{Epochs} & \textbf{Hidden units}  & \textbf{Latent dim.}   & \textbf{N samples (train)} & \textbf{N samples (test)}\\ 
\hline
 local\_cl$i$ & 1 & $150\times10$ & \multirow{3}{*}{256} & \multirow{3}{*}{20} & \multirow{3}{*}{50} & \multirow{3}{*}{10000} \\\cline{1-3}
centralized & 1 & $150\times10\times C$ & & & & \\\cline{1-3}
federated & 150 & 10 & & & &
\end{tabular}
\end{table}


\begin{table}[h!]
\caption{ADNI data, MAR setting, Natural split. Testing dataset.}
\label{tab:natural_MAR_test}
\centering
\begin{tabular}{p{1.8cm}||g|g|l|l|l}
\textbf{Training} & \multicolumn{2}{G|}{Federated} & \multicolumn{3}{c}{Centralized} \\
\hline
\textbf{Model} & {\textbf{FedProx}} & FedProx\_loc & mean & ice & \textbf{random forests} \\
\hline
\textbf{MSE $\boldsymbol{\downarrow}$ \mbox{mean (std)}} & \textbf{0.4458 (0.0083)} & 0.5086 (0.0383) & 1.117 (0.0555) & 1.294 (0.2165) & \textbf{0.424 (0.0338)}
\end{tabular}
\end{table}


\begin{table}[h!]
\caption{ADNI data, MCAR setting, not-IID split. Testing dataset.}
\label{tab:notIID_MCAR_test}
\centering
\begin{tabular}{p{1.8cm}||g|g|l|l|l}
\textbf{Training} & \multicolumn{2}{G|}{\textbf{Federated}} & \multicolumn{3}{c}{Centralized} \\
\hline
\textbf{Model} & {\textbf{FedProx}} & FedProx\_loc & mean & ice & random forests \\
\hline
\textbf{MSE $\boldsymbol{\downarrow}$ \mbox{mean (std)}} & \textbf{0.4477 (0.0053)} & 0.627 (0.008) & 1.1717 & 0.7157 (0.006) & 0.5167 (0.0009)
\end{tabular}
\end{table}

\end{document}